\pdfoutput=1

\documentclass[11pt]{article}

\usepackage{acl}

\usepackage{times}
\usepackage{latexsym}
\usepackage{graphics}
\usepackage{graphicx}

\usepackage[T1]{fontenc}

\usepackage[utf8]{inputenc}

\usepackage{microtype}
\newcommand{\rd}[1]{{\color{red} #1}}
\newcommand{\bl}[1]{{\color{blue} #1}}
\newcommand{\RNum}[1]{\uppercase\expandafter{\romannumeral #1\relax}}

\title{The Secret of Metaphor on Expressing Stronger Emotion}

\author{Yucheng Li~\textsuperscript{1}, Frank Guerin~\textsuperscript{1},Chenghua Lin~\textsuperscript{2}\thanks{~~Corresponding author}  \\
\textsuperscript{1}~Department of Computer Science, University of Surrey, UK \\
\texttt{\{yucheng.li,f.guerin\}@surrey.ac.uk}\\ \textsuperscript{2}~Department of Computer Science, University of Sheffield, UK \\
\texttt{c.lin@sheffild.ac.uk }
}

\begin{document}
\maketitle
\begin{abstract}

Metaphors are proven to have stronger emotional impact than literal expressions. Although this conclusion is shown to be promising in benefiting various NLP applications, the reasons behind this phenomenon are not well studied. 
This paper conducts the first study in exploring how metaphors convey stronger emotion than their literal counterparts. We find that metaphors are generally more specific than literal expressions. The more specific property of metaphor can be one of the reasons for metaphors' superiority in emotion expression. When we compare metaphors with literal expressions with the same specificity level, the gap of emotion expressing ability between both reduces significantly. In addition, we observe specificity is crucial in literal language as well, as literal language  can express stronger emotion by making it more specific.

\end{abstract}

\section{Introduction}

Metaphors are widely used in human language, which allows people to communicate not just information, but also feelings and attitudes. It is generally believed that metaphors are especially effective in expressing subjective elements, such as sentiment and attitude. 
Recent studies in Psychology and Computational Linguistics thus provide a wide range of qualitative evidence which supports the idea that metaphors are closely related to sentiment.
For example, \citet{rentoumi2012-meta-in-sentiment} use metaphorical expressions as a feature in sentiment polarity detection and find it can be an effective indicator. \citet{mao2021-multitask-meta-absa} introduce a multitask framework which jointly optimizes a metaphor detection task and aspect-based sentiment analysis and observe considerable improvement on both tasks. More importantly, \citet{moh-16} give the first quantitative finding which shows that 83.6\% of annotated metaphors tend to have a stronger emotional impact than their literal counterparts.

However, although researchers conduct fruitful studies showing how metaphors are closely related to sentiment, the reason behind this phenomenon is not well explored. Investigating the mechanism of metaphor sentiment interaction can be quite promising. For instance, understanding how metaphor builds emotional bonds can guide metaphor generation models \cite{metagen,li-2022-cm-gen} producing empathetic and persuading responses. The result can also be helpful for sentiment analysis, especially on metaphor-enriched text \cite{cabot2020-meta-emotion-interaction-in-politic}.

\begin{figure}
    \centering
    \includegraphics[width=\columnwidth]{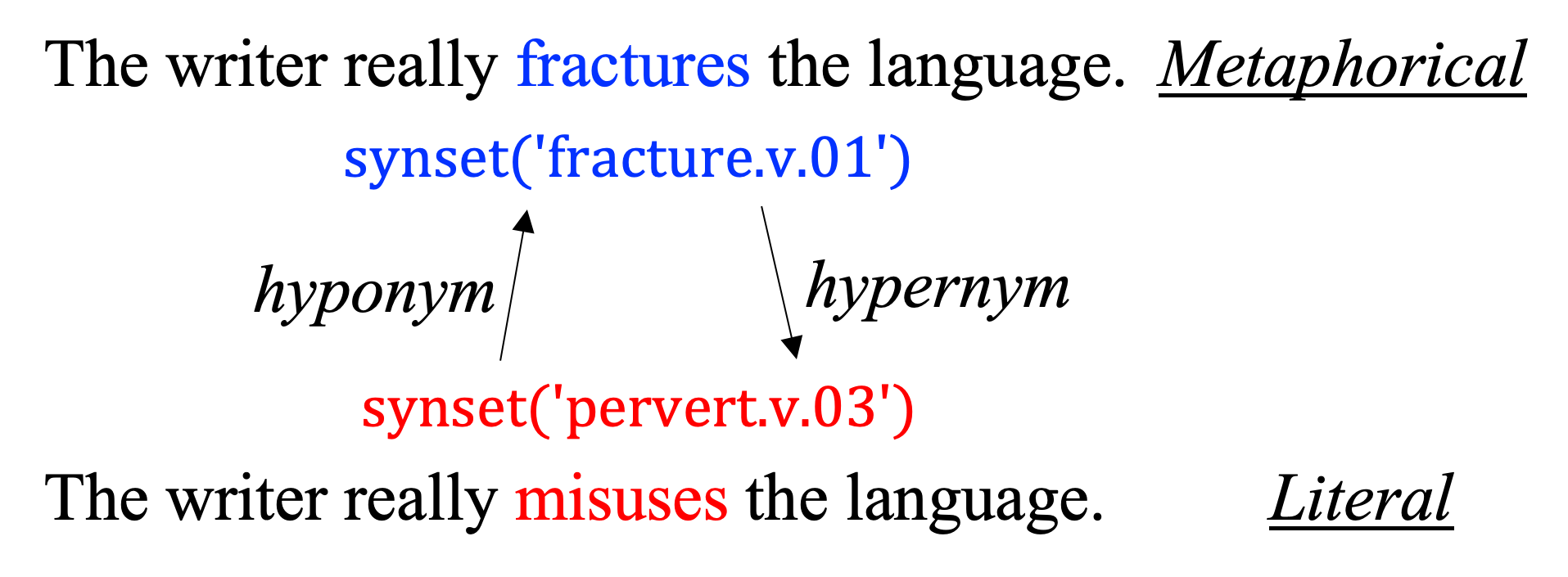}
    \caption{\textit{hypernym} and \textit{hyponym} relation between metaphor and literal expressions. Synset here presents the word sense of the target word based on WordNet sense dictionary. Blue text indicates metaphor and red text indicates literal.}
    \label{fig:example}
    \vspace{-0.5em}
\end{figure}

\begin{figure*}[ht]
    \centering
    \includegraphics[width=\textwidth]{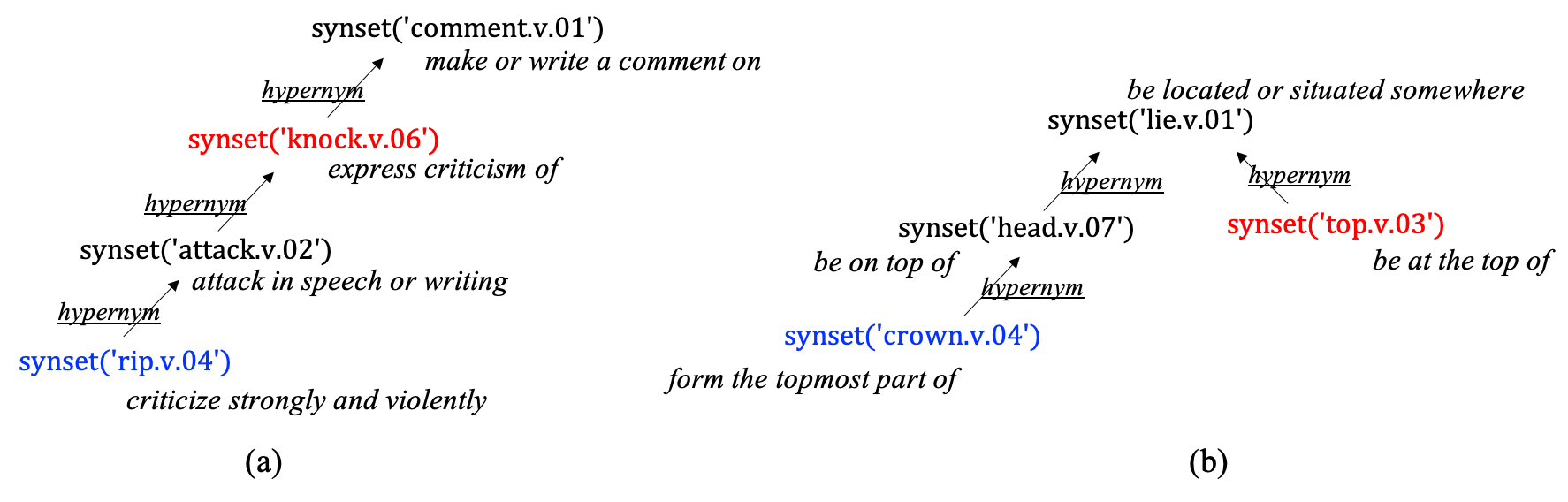}
    \caption{Two cases illustrating positions of metaphorical and literal synset in the WordNet hierarchy.}
    \label{fig:hierarchy}
\end{figure*}

In this paper, we introduce an exploratory answer to the question of how metaphors convey stronger emotion than literal language. To investigate this phenomenon, we manually analyse the metaphor-literal parallel corpus from MOH dataset \cite[see example in Figure \ref{fig:example}]{moh-16} where the more emotional expression is marked among each metaphor-literal pair. Our study finds that metaphors might impose emotional impact on readers via giving more specific expressions, i.e., making the expression more precise. First, we find most metaphorical expressions are more specific than their literal counterparts. In other words, literal translations of metaphors usually convey more general meanings. It suits our intuition that metaphors are believed as more vivid. Second, we find metaphor's stronger emotional impact is partially from its more specific description. When we compare metaphors and their literal counterparts where both share the same level of specificity, we find the superiority of metaphors in arousing emotional impact drops significantly. When we test the \textit{more-specific} principle on literal expressions, we find more specific literal expressions do surpass general ones on emotional impact.

We use linguistic relation \textit{hypernym} and \textit{hyponym} from WordNet \cite{miller1995wordnet} to define the specificity in our analysis. Specifically, \textit{hypernym} denotes a word with a broad meaning yet \textit{hyponym} denotes a word with a more specific meaning. So if a literal term is its metaphorical counterpart's \textit{direct hypernym}, we know the metaphor describes a more specific meaning, or to say in a metaphorical way, draws a more precise picture. The Figure \ref{fig:example} shows an example of the above situation: the synset of the literal expression \textit{misuses language} is the direct hypernym of the metaphorical synset, which means the literal expression is more general and the metaphor is more specific.

In case there is no direct hypernym or hyponym relation between metaphorical and literal expression, we compare the place of both in the WordNet Hierarchy to determine which one is more specific. In Figure \ref{fig:hierarchy}, we see clearly that from the top to the bottom in the WordNet hierarchy, expressions tend to be more specific. So we can determine the relative specificity of terms by comparing their relative position in the hierarchy.

In summary, our contributions are mainly in two folds: 1) we introduce a novel hypernym-hierarchy method to measure the specificity of language expression and find metaphors are usually more specific than literal counterparts; 2) we find the reason why metaphor express stronger emotion is partially due to its more specific expression. Our code and data can be found in \url{https://github.com/liyucheng09/Metaphors_are_more_emotional}

\section{The MOH Dataset}

\citet{moh-16} create a metaphor dataset in which verb senses are annotated for both metaphoricity and emotionality. In addition, the metaphorical uses are paired with their human-validated interpretations in the form of literal paraphrases (i.e., the metaphor's literal counterpart). In Table \ref{tab:moh_case}, we give an example of the MOH annotation for the verb \textit{rip}. There are 171 metaphor-literal parallel annotations in total. We employ the MOH dataset in our study due to its parallel feature.











\begin{table}[t]
    \centering
    \resizebox{\columnwidth}{!}{
    \begin{tabular}{l|p{6cm}}
    \hline
       \textbf{Term:}  & rip \\
       \textbf{Sense/Synset:}  &  Synset(`rip.v.04') \\
       \textbf{Sentences:} & The candidate \bl{ripped} into his opponent mercilessly. \\
       \textbf{Literal:} & The candidate \rd{criticized}  his opponent mercilessly. \\
       \textbf{Emotion:} & The metaphorical expression is more emotional. \\ \hline
    \end{tabular}}
    \caption{The annotation example of verb \textit{rip} in the MOH dataset.}
    \label{tab:moh_case}
    \vspace{-0.5em}
\end{table}

\section{Experimental Setup}

This study tests two research hypotheses:

\begin{quote}
    \textit{Hypothesis 1}: Metaphors are generally more specific than their literal counterparts. In other words, metaphors are lower than their literal counterparts in the WordNet hierarchy.
    
    \textit{Hypothesis 2}: Metaphors' stronger emotional impact is partially from metaphors' more specific expression. In other words, more precise expression is one of the reasons why metaphors convey stronger sentiment than their literal counterparts.
\end{quote}

To compare the specificity of metaphor and its literal counterpart, the hypernym-hierarchy information is assigned to both in parallel. 

To explore the role specificity plays in the interaction between emotion and metaphor, we first analyse the correlation between specificity and emotion label of metaphors. We then perform two more experiments to further test how specificity affects emotional impact: 1) labelling which one is more emotional between metaphor and literal counterpart with the same level of specificity; 2) labelling which one is more emotional between a more general literal expression and a more specific literal expression. The first test isolates the influence of specificity in the emotion comparison of metaphor-literal pair; the second  tests whether specificity empowers literal expression to convey stronger sentiment.

\begin{table*}[t]
    \centering
    \resizebox{\textwidth}{!}{
    \begin{tabular}{lp{5cm}lp{5cm}ll}
    \hline
    Term 1 & Sentence 1 & Term 2 & sentence 2 & Specific & Emotion \\ \hline
    Synset(rip.v.04) & The candidate \bl{ripped} into his opponent mercilessly. & Synset(criticize.v.01) & The candidate \rd{criticized} into his opponent mercilessly. & first & first \\
    Synset(rip.v.04) & The candidate \bl{ripped} into his opponent mercilessly. & Synset(barrage.v.01) & The candidate \rd{admonished} his opponent mercilessly. & same & same \\
    Synset(criticize.v.01) & The candidate \rd{criticized}  his opponent mercilessly. & Synset(attack.v.02) & The candidate \rd{scolded} his opponent mercilessly. & second & second \\ \hline
    \end{tabular}}
    \caption{Examples of sentence pairs in three experiments. The specific column denotes which sentence if more specific, and the emotion column indicates which sentence is more emotional. Blue text is metaphor and red text is literal. The three examples are  \textit{metaphor vs. literal, metaphor vs. specific literal, and literal vs. more specific literal} respectively.}
    \label{tab:parallel_cases}
    \vspace{-0.6em}
\end{table*}

\subsection{Specificity Test}
\label{specificity_test}

\noindent\textbf{Synset annotation. }
To access the hypernym relation of metaphor-literal data or locate both in the WordNet hierarchy, synsets of both need to be annotated. A synset in WordNet can be seen as a word sense item thus annotating a synset can be regarded as a word sense disambiguation task. The overall annotation procedure is as follows: 1) query WordNet with lemmatized target words to obtain synsets candidates; 2) determine the best suiting synset for both metaphorical and literal targets based on synset gloss and example sentence. An example of synset annotation is in Figure \ref{fig:example}, where target words (i.e., metaphor and its literal counterpart, in colour) are labelled with synset.

\noindent\textbf{Determining Specificity. } After obtaining the synsets of metaphor-literal pair, there are two ways to determine the relative specificity of both expressions. For cases where there is a \textit{direct hypernym} or \textit{direct hyponym} relation between metaphorical and literal synset, we can know the relative specificity explicitly: the hypernym is more general yet the hyponym is more specific. For cases where metaphorical and literal synset are not connected with such a relation, we locate both terms in the WordNet hierarchy and compare their relative position. The locating procedure is as follows: 1) find their lowest common hypernym in WordNet hierarchy; 2) compute the number of hops from their common hypernym to both terms.

The example shown in Figure \ref{fig:example} belongs to the first situation that is there is a direct relation linking the two terms. So does the Figure \ref{fig:hierarchy} \texttt{(a)} case, where the literal term and the metaphoric term are connected via two hops of hypernym relations. So we know the literal term, as it is the hypernym of the metaphoric term, is more general than its metaphoric counterpart.
In contrast, examples in Figure \ref{fig:hierarchy} \texttt{(b)} is the second situation, where the lowest common hypernym has to be found. It takes two hops from the metaphorical synset to reach the common nearest common hypernym, but it only takes one hop for its literal counterpart to arrive at the common hypernym. So we know the lower synset (i.e., the metaphorical one) is relatively more specific than the other. In our experiments, we find the first situation is the dominant cases, which suits around 86\% (98 out of 114) of metaphor-literal pairs we tested. And only 14\% (16 out of 114) pairs fall in the second situation.

\subsection{Emotional Impact Test}

To investigate the emotional impact that comes from the specificity of expressions, we analyse the correlation between the specificity and emotion label of metaphors. To further explore the interaction between specificity and emotional impact, we conduct two more manual experiments. 

\textbf{First}, we compare which is more emotional between metaphor and its literal counterparts with the same level of specificity. To perform the comparison, we need to make up literal paraphrase same specific as the metaphor. We use the \textit{sister terms} relation in WordNet to realise it. Two terms are sister terms as long as they share the same hypernym in WordNet, which means sister terms are at the same level in the WordNet hierarchy. We manually choose an appropriate literal sister term of the metaphor, and paraphrase the origin sentence with the literal term to form a literal counterpart has the same level of specificity. See line 2 in Table \ref{tab:parallel_cases} for an example of such a sentence pair. With the paired data, we employ three human annotators with linguistics backgrounds to judge which expression is more emotional.

\textbf{Second}, we compare which is more emotional between more general literal and more specific literal expression. We use the \textit{direct hyponym} relation to realise it. Similarly, we manually choose a direct hyponym term of the literal expression and paraphrase the origin sentence with the more specific literal term to make up the more specific counterpart. See line 3 in Table \ref{tab:parallel_cases} for such a example sentence pair. We invite the same three annotators to tackle the emotion annotation, where annotators have to decide which express is more emotional, or choose the third option saying that both are similarly emotional.

\section{Results}

\subsection{Metaphor and Specificity}

We obtain 114 valid metaphor-literal pairs in the specificity experiment. 54 instances are invalid because we find no common hypernym among the metaphorical and literal terms in WordNet hierarchy. Among all 114 valid cases, 78.9\% of metaphors are lower than their literal counterparts in the WordNet hierarchy, which means they are generally more specific. Only 5.2\% of pairs show the opposite result, which means the metaphors are more general. 15.7\% of metaphor-literal pairs are at the same specificity level. So in summary, we present a quantitative result that shows metaphors are generally more specific than literal expressions.
Perhaps that is the reason why metaphors are believed giving more vivid descriptions. 

\subsection{Specificity and Emotional Impact}
\label{level_and_emotion_result}

\noindent\textbf{Metaphor Specificity and Emotion. }
Based on both emotion and specificity labels of metaphor-literal pairs, we measure the correlation between these two dimensions. The results are shown in Table \ref{tab:meta_spe_emo}. According to the table, we find that specificity can be a strong indicator of the emotional impact. 
Among all 90 more specific metaphors, 91.1\% of them express stronger emotion. From the emotional dimension, 84.5\% of metaphors that express stronger emotion are also more specific than their literal counterparts. 

\begin{table}[t]
    \centering
    \resizebox{\columnwidth}{!}{
    \begin{tabular}{l|ccc}
    \hline
    Metaphors are .. & more specific & more general & same \\ \hline
        more emo. & 82 (71.9\%) & 10 (8.7\%) & 5 (4.4\%) \\
        less/same emo. & 8 (7.0\%) & 8 (7.0\%) & 1 (0.8\%) \\
    \hline
    \end{tabular}
    }
    \caption{
    When metaphors are more specific (general) than literal expressions, will they be more (less) emotional at the same time?}
    \label{tab:meta_spe_emo}
    \vspace{-0.8em}
\end{table}

\begin{table}[t]
    \centering
    \resizebox{\columnwidth}{!}{
    \begin{tabular}{l|ll}
    \hline
         Metaphors are ... & vs. Literal & vs. Specific Literal \\ \hline
        more emo. & 143 (83.6\%) & 42 (40.0\%, $\downarrow$ \textbf{43.6}\%)\\
        less emo. & 17 (9.9\%) & 23 (21.9\%, $\uparrow$ 12.0\%)\\
        similarly emo. & 11 (6.4\%) & 40 (38.1\%, $\uparrow$ \textbf{31.7}\%) \\
        Total & 171 & 105\\
        \hline
    \end{tabular}}
    \caption{
    Which is more emotional, metaphor or literal? Comparisons made between metaphors vs. normal literals and metaphors vs. more specific literals.}
    \label{tab:meta_literal_more_spe_literal}
\end{table}

\noindent\textbf{Metaphor and More Specific Literal. }
To investigate the extent to which specificity influences the emotional impact of metaphors, we perform an experiment to compare metaphors with general literal expressions and literal expressions sharing the same level of specificity with metaphors. We construct 105 valid sentence pairs in total. We fail to make up more because we cannot find a literal synset with the same specificity of the metaphor for those cases. The results are presented in Table \ref{tab:meta_literal_more_spe_literal}. The inner-annotator agreement (IAA) score for emotion labelling is 0.77 via Krippendorff’s alpha \cite{krippendorff2011-iaa}. The first column of the Table is obtained from MOH's result. We find that the superiority of metaphors in expressing sentiment drops significantly from 83.6\% to 40.0\% when metaphors are compared to more specific literal expressions. In contrast, when metaphor-literal pairs share the same specificity, the ratio of expressing similar emotional strength increases noticeably. This results show that specificity is clearly a factor associating with emotional strength. However, metaphors still tend to have more emotional impact than more specific literal expressions. So we believe there are more factors affecting sentiment expressing ability despite specificity. We leave it to future works.

\begin{table}[t]
    \centering
    \resizebox{\columnwidth}{!}{
    \begin{tabular}{lr}
    \hline
        \# instances that are: \\
        \qquad more specific is more emotional & 32 (34.8\%) \\
        \qquad more general is more emotional & 14 (15.2\%) \\
        \qquad similarly emotional & 46 (50.0\%) \\
        \qquad Total & 92 \\
    \hline
    \end{tabular}
    }
    \caption{Which is more emotional, literals or more specific literals?}
    \label{tab:literal_more_spe_literal}
    \vspace{-0.8em}
\end{table}

\noindent\textbf{Literal and More Specific Literal. }
To test whether the \textit{more-specific} mechanism also applies to literal expressions, we compare literal expressions with more specific literal ones. We construct 92 such sentence pairs in total. The IAA score of emotion labelling in this experiment is 0.82. The results are shown in Table \ref{tab:literal_more_spe_literal}, which illustrate that more specific expressions do impose a stronger emotional impact than more general ones. This demonstrates that specificity can be a stronger indicator in sentiment analysis in both figurative language and literal language.



\bibliography{master}
\bibliographystyle{acl_natbib}




\end{document}